\documentclass{article}
\usepackage{spconf,amsmath,graphicx}
\usepackage{subcaption}
\usepackage{booktabs}
\usepackage{multirow}
\usepackage{xcolor}
\usepackage{comment}
\usepackage{hyperref}

\def\adam/{Adam}

\def\asru/{ASR-U}
\def\fairseq/{fairseq}
\def\nvidia/{GTX 1080Ti}
\def\eodm/{EODM}
\def\kmeans/{$K$-means}
\def\wavtovecu/{wav2vec-U}
\def\wavtovectwo/{wav2vec 2.0}
\def\skipgram/{skipgram}
\def\skipgrams/{skipgrams}
\def\readout/{wav2vec 2.0 readout}
\def\unibi/{uni+bi-grams}
\def\unibitri/{uni+bi+tri-grams}
\def\unibifour/{uni+bi+4-grams}
\def\unifour/{uni+4-grams}
\def\unifive/{uni+5-grams}

\usepackage{amsmath,amsfonts,bm}

\def\Figref#1{Fig.~\ref{#1}}

\def\eqref#1{Eq.~\ref{#1}}
\def\Eqref#1{Eq.~(\ref{#1})}

\def\1{\bm{1}}

\def\rvx{{\mathbf{x}}}
\def\rvy{{\mathbf{y}}}

\def\rmA{{\mathbf{A}}}

\DeclareMathAlphabet{\mathsfit}{\encodingdefault}{\sfdefault}{m}{sl}
\SetMathAlphabet{\mathsfit}{bold}{\encodingdefault}{\sfdefault}{bx}{n}

\def\gK{{\mathcal{K}}}

\def\gX{{\mathcal{X}}}
\def\gY{{\mathcal{Y}}}

\def\sX{{\mathbb{X}}}
\def\sY{{\mathbb{Y}}}

\newcommand{\Ls}{\mathcal{L}}

\newcommand{\sigmoid}{\sigma}

\title{UNSUPERVISED SPEECH RECOGNITION WITH N-SKIPGRAM AND POSITIONAL UNIGRAM MATCHING}
\name{Liming Wang$^{\star}$, Mark Hasegawa-Johnson$^{\star}$ and Chang D. Yoo$^{\dagger}$}
\address{$^{\star}$Department of Electrical and Computer Engineering, University of Illinois Urbana-Champaign \\ $^{\dagger}$Artificial Intelligence and Machine Learning Lab, KAIST}
\begin{document}
%
\maketitle
\begin{abstract}
Training unsupervised speech recognition systems presents challenges due to GAN-associated instability, misalignment between speech and text, and significant memory demands. To tackle these challenges, we introduce a novel ASR system, ESPUM. This system harnesses the power of lower-order $N$-skipgrams (up to $N=3$) combined with positional unigram statistics gathered from a small batch of samples. Evaluated on the TIMIT benchmark, our model showcases competitive performance in ASR and phoneme segmentation tasks. Access our publicly available code at \href{https://github.com/lwang114/GraphUnsupASR}{https://github.com/lwang114/GraphUnsupASR}.
\end{abstract}
\section{Introduction}
\label{sec:intro}
Learning a speech recognizer with only unpaired speech and text corpora, or unsupervised speech recognition (\asru/)\cite{Glass2012-unsup-speech}, is a self-supervised learning task crucial for developing speech technology for low-resource languages. Beyond converting speech to text without reliance on transcribed speech, an \asru/ system can serve as the linchpin for low-resource text-to-speech synthesis~\cite{Ni-etal-2022-unsup-tts,Liu2022-ttsu}, speech translation~\cite{Wang2023-unsup-mt,Fu2023-unsup-mt} and other spoken language understanding tasks.
Despite significant strides made in the domain~\cite{Liu2018-asru,Chen2019-asru,Yeh2019-asru,Baevski2021-wav2vec-u,Liu2022-asru}, the \emph{stability} of \asru/ systems remains a conspicuous bottleneck~\cite{Lin2022-asru-robustness,Ni-etal-2022-unsup-tts}. Many leading \asru/ models, including the current state-of-the-art models, \wavtovecu/ and its 2.0 iteration~\cite{Baevski2021-wav2vec-u,Liu2022-asru}, rely heavily on generative adversarial networks (GAN)~\cite{Goodfellow2014}. These GANs are notoriously difficult to train, demanding rigorous regularization and hyperparameter tuning, often displaying sensitivity to the relative weightings of regularization losses ~\cite{Ni-etal-2022-unsup-tts}. 

The only existing method that bypasses the need for GANs is the empirical output distribution matching (\eodm/) approach \cite{Yeh2019-asru}. This approach trains an ASR system to directly match the empirical $N$-gram distribution of authentic phoneme sequences, eliminating the need for a discriminator. However, straightforward $N$-gram matching encounters difficulties: the quantity of unique $N$-grams quickly grows to intractable number and the accuracy of approximating the $N$-gram distribution diminishes as $N$ increases. 
Notably, \cite{Yeh2019-asru} revealed the necessity of $5$-gram and large batch size (50,000 tokens per batch) for \eodm/ to yield optimal \asru/ outcomes. Due to memory restrictions, they consider only the top 10,000 $5$-gram distributions.

An additional pivotal challenge for \asru/ lies in phoneme segmentation. While \cite{Chen2019-asru} utilized a fixed phoneme segmentation derived from an unsupervised model~\cite{Wang2017-unsup-seg} during GAN training, \cite{Yeh2019-asru} opted for a block-wise alternative minimization technique to refine segmentations. By contrast, \wavtovecu/ (2.0)~\cite{Baevski2021-wav2vec-u,Liu2022-asru} performs segmentation by directly merging successive frames assigned to identical phonemes. Unfortunately, current methodologies have yet to incorporate the innovative strides made in unsupervised phoneme segmentation~\cite{Kreuk2020-cpc-seg,Bhati2021-segmental-cpc,Strgar2021-readout-seg}. These recent techniques improve the quality of the detected phoneme boundaries by employing differentiable self-supervised learning objectives, such as (segmental) contrastive predictive coding (CPC)~\cite{Kreuk2020-cpc-seg,Bhati2021-segmental-cpc} and teacher-student learning~\cite{Strgar2021-readout-seg}.

In this paper, we proposed Empirical Skipgrams and Positional Unigram Matching (ESPUM), a novel GAN-free \asru/ model based on $N$-\skipgrams/ and positional unigram matching. Our model achieves competitve \asru/ performance on the standard TIMIT~\cite{Garofolo1993} benchmark while being more consistent in different hyperparameter settings than the GAN-based approach. Further, our model is more memory-efficient than the previous GAN-free \asru/ models~\cite{Yeh2019-asru} by requiring only lower-order $N$-\skipgrams/ (up to $N=3$) and positional unigram information. Last but not least, we design a novel differentiable phoneme segmenter end-to-end trainable with the rest of the \asru/ systems and outperform all previous methods in the unsupervised phoneme segmentation task.

\section{Problem formulation}
\label{sec:formulation}
Suppose we have an unlabeled speech corpus consisting of speech feature sequences $X^{(i)}\in \gX^T \sim P_X,i=1,\cdots,n_X$ and another unpaired text-only corpus containing phoneme label sequences $Y^{(j)}\in \gY^L\sim P_Y,j=1,\cdots,n_Y$. Suppose that the speech and text corpora are \emph{matched} in the sense that there exists a speech recognizer (ASR) $G:\sX^T\mapsto \sY^L$ such that for any phoneme sequence $\forall \rvy\in \sY^L$,
\begin{align}\label{eq:asru}
    P_X\circ G(\rvy) := \sum_{\rvx\in \sX^T}P_X(\rvx)G_{\rvy}(\rvx) = P_Y(\rvy).  
\end{align}
The goal of \asru/ is to recover such an ASR.

\section{Method}
\begin{figure}
    \begin{subfigure}{0.5\textwidth}
        \centering
        \includegraphics[width=1.0\textwidth]{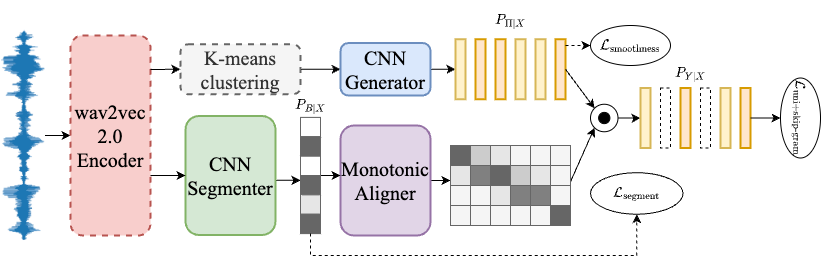}
        \caption{Overall architecture of the proposed model. Dashed components are not trainable}
        \label{fig:skipgram_matcher}
    \end{subfigure}
    \begin{subfigure}{0.5\textwidth}
        \centering
        \includegraphics[width=1.0\textwidth]{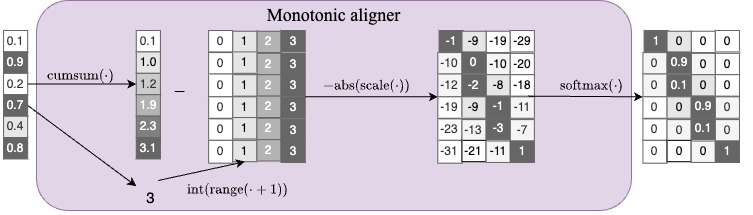}
        \caption{Differentiable monotonic aligner}
        \label{fig:monotonic_aligner}
    \end{subfigure}
\end{figure}
While most existing methods employ a GAN~\cite{Chen2019-asru,Baevski2021-wav2vec-u} for learning such an ASR, we propose a GAN-free method consisting of five main modules as shown in \Figref{fig:skipgram_matcher}.
\subsection{Self-supervised speech generator}
Our model first takes raw speech waveform as input to a pretrained, \emph{self-supervised speech encoder} such as \wavtovectwo/~\cite{Baevski2020-wav2vec2}. Empirically we found that the raw features lead to unstable training, and instead we use a \emph{\kmeans/ clustering} module to discretize the speech features into one-hot vectors. A convolutional neural network (CNN) \emph{generator} (the ASR) then converts the one-hot vectors into phoneme probabilities.

\subsection{CNN segmenter}
While solving \Eqref{eq:asru} over all possible ASR is both statistically and computationally infeasible, in practice we can constrain the structure of the ASR significantly if we have access to \emph{noisy, unsupervised} phoneme-level segmentations $\tilde{B}_{1:T}^{(i)},i=1,\cdots,n_X$, where $\tilde{B}_t=1$ if frame $t$ is a boundary between two phonemes and otherwise $0$ if it is within a phoneme. Using this labels, we train a  CNN-based \emph{segmenter} to predict the phoneme boundaries from speech features: 
\begin{align}
    P_{\theta}[B_t=1|X_{1:T}=\rvx] := \sigmoid(\texttt{CNN}_{\theta}(\rvx)),
\end{align}
where $\sigma(\cdot)$ is the sigmoid function. The segmenter is then trained using a weighted binary cross-entropy (BCE) loss $\Ls_{\mathrm{segment}}(\theta)$. 

\subsection{Monotonic alignment}
Due to the noise in the segment labels, the raw labels from the segmenter often lead to severe misalignments between the speech features and phoneme labels. To address this issue, we propose a ``soft'' \emph{monotonic alignment} $\rmA^\theta \in [0,1]^{L\times T}$ from the segmenter by a sequence of differentiable operation shown in \Figref{fig:monotonic_aligner}, inspired by a similar mechanism from the Segmental CPC model~\cite{Bhati2021-segmental-cpc}. Using this alignment, we then compute a sequence of segment-level features from the frame-level features as
\begin{align}
    \bar{X}_l^\theta = \sum_{t=1}^T A_{lt}^\theta X_{t}.
\end{align}  
Note that this soft monotonic alignment can be trained end-to-end with the speech generator with the ASR-U criteria, allowing the model to refine the segment boundaries using information from the unpaired text data.

\subsection{$N$-skipgram and positional unigram matching}
After the monotonic alignment, the ASR now takes a much simpler form as it can predict each phoneme label independently given each segment-level speech feature. We can then learn the ASR by performing distribution matching between the segmented speech distribution $P_{\bar{X}_{1:L}}^\theta$ and the text distribution $P_{\bar{Y}_{1:L}}$. Instead of matching the full distributions or $N$-grams, we find it much more memory-efficient and reliable to use (bi-)\emph{skipgrams} and more generally, $N$-skipgrams defined for skip sizes $\mathbf{k} := (k_1,\cdots,k_{N-1})$ as
\begin{align}\label{eq:N_skipgram}    
P_{Y,\mathbf{k}}(y_{1:N}) := P_{Y_1,Y_{1+k_1},\cdots,Y_{1+\sum_{l=1}^{N-1}k_l}}(y_{1:N}).
\end{align}
Using \Eqref{eq:asru}, we then learn the ASR $G$ by minimizing the following matching loss:
\begin{align}\label{eq:skipgram_loss}
    \Ls_{\mathrm{skipgram}}(G, \theta) := \sum_{\mathbf{k}\in \gK}\left\|\hat{P}_{Y, \mathbf{k}}- \hat{P}_{\bar{X}, \mathbf{k}}^{\theta}\circ G\right\|_1,
\end{align}
where $\gK$ is the set of skip sizes used and $\hat{P}_{Y,\mathbf{k}}$ and $\hat{P}^{\theta}_{\bar{X},\mathbf{k}}$ are empirical distributions estimated from sample batches. Further, inspired by \cite{Bhati2021-segmental-cpc}, we also use the \emph{positional unigram} $P_{\bar{X}_l}^\theta$ and $P_{Y_l}$ by another $L_1$ loss:
\begin{align}\label{eq:unigram_loss}
    \Ls_{\mathrm{unigram}}(G, \theta) := \sum_{l=1}^L \|\hat{P}_{Y_l}-\hat{P}_{X_l}G\|_1.
\end{align}

\subsection{Smoothness and segment relabeling}
Similar to previous works~\cite{Yeh2019-asru,Baevski2021-wav2vec-u}. we also also apply the smoothness loss to encourage similar phoneme labels for nearby speech feature frames:
\begin{align}\label{eq:smooth}
    \Ls_{\mathrm{smooth}}(G) := \sum_{i=1}^{n_X}\sum_{t=1}^T \left\|G(X_{t+1}^{(i)}) - G(X_t^{(i)})\right\|_2^2.
\end{align}
The overall training objective is then
\begin{align}\label{eq:overall}
    \Ls_{\mathrm{unigram}}+\Ls_{\mathrm{skipgram}} + \Ls_{\mathrm{segment}} + \lambda_{\mathrm{smooth}}\Ls_{\mathrm{smooth}}.
\end{align}

Moreover, to further improve the segmentation quality, we replace the older, noisier labels $\tilde{B}_{1:T}$ with the predicted labels from the segmenter $\tilde{B}'_{1:T}$ after training converges using the older labels, a process called \emph{segment relabeling}.

\section{Experiments}
\begin{table}[t]
    \centering
    \resizebox{0.45\textwidth}{!}{
    \begin{tabular}{l|c|c|cccc}
    \toprule
    & \multirow{2}{*}{GAN?} & \multirow{2}{*}{LM} & \multicolumn{2}{c}{matched} & \multicolumn{2}{c}{unmatched}\\
    \cmidrule(lr){4-5}\cmidrule(lr){6-7}
    & & & val & test & val & test \\
   \midrule
   EODM~\cite{Yeh2019-asru} & \multirow{2}{*}{No} & \multirow{2}{*}{5-grams} & - & 42.6 & - & 49.1 \\
    + HMM ST & & & - & 36.5 & - & 41.6 \\
   \wavtovecu/~\cite{Baevski2021-wav2vec-u} & \multirow{2}{*}{Yes}  & \multirow{2}{*}{4-grams}  & 17.0 & 17.8 & 21.3 & 22.3 \\
    + CTC ST & & & 11.3 & 12.0 & 13.8 & 15.0\\
    \midrule
        \multicolumn{7}{c}{\emph{Proposed models}}\\
    \midrule
    uni+bi+tri & \multirow{3}{*}{No} & \multirow{3}{*}{4-grams} & 42.9 & 43.3 & 51.3 & 47.3\\
    uni+bi+tri, iter 1 & & & 39.4 & 39.1 & 49.8 & 45.1 \\
    + HMM ST & & & 33.1 & 33.7 & 47.0 & 42.9\\
    \bottomrule
    \end{tabular}
    }
    \caption{PER of various \asru/ models on TIMIT.}
    \label{tab:asru_timit}
\end{table}

\begin{table}[t]
    \centering
    \resizebox{0.45\textwidth}{!}{
    \begin{tabular}{lcccccccc}
    \toprule
    & \multicolumn{2}{c}{P}  & \multicolumn{2}{c}{R}  & \multicolumn{2}{c}{F1} & \multicolumn{2}{c}{R-val}\\ 
    \cmidrule(lr){2-3}\cmidrule(lr){4-5}\cmidrule(lr){6-7}\cmidrule(lr){8-9}
    & H & L & H & L& H & L & H & L \\
   \midrule
    \multicolumn{9}{c}{\emph{Speech-only}}\\
   \midrule
   Kreuk et al~\cite{Kreuk2020-cpc-seg} & 81.4 & 85.3  & 76.5 & 83.5 & 78.9 & 84.4  & 81.7 & 86.6 \\
   Kreuk et al~\cite{Kreuk2020-cpc-seg}$^*$ & 78.3	& 85.8 & 75.8 &	82.7 & 77.1	& 84.2 & 80.4 & 86.3 \\
   Bhati et al~\cite{Bhati2021-segmental-cpc} & - & 84.6 & -  & 86.0 & - & 85.3 &  - & 87.4 \\
   Strgar et al~\cite{Strgar2021-readout-seg} & 82.4 & 91.0 & 81.2 & 88.5 &  81.8 & 89.7 & 84.5 & 91.0 \\
   Strgar et al~\cite{Strgar2021-readout-seg}$^*$ & 82.6 & 89.6 & 74.8 & 81.6 & 78.5 & 85.4 & 81.0 &  86.4 \\
    \midrule
    \multicolumn{9}{c}{\emph{Speech+unpaired text}}\\
    \midrule
    EODM~\cite{Yeh2019-asru} & - & 80.9 & - & 84.3 & - & 82.6 & - & 84.8\\
    wav2vec-U+HMM ST~\cite{Baevski2021-wav2vec-u} & 67.8 & 74.3 & 74.4 & 80.0 & 71.0 & 77.1 & 73.8 & 79.5 \\
    Ours (matched) & 88.9 & 93.3 & 77.3 & 83.9 & 82.7 & 88.4 & 83.5 & 88.4 \\
    Ours (matched, iter 1) & \textbf{87.2} & \textbf{93.4} & \textbf{85.3} & \textbf{89.3} & \textbf{86.2} & \textbf{91.3} & \textbf{88.1} & \textbf{91.9} \\ 
    Ours (unmatched, iter 1) & 88.2 & 90.8 & 76.4 & 84.0 & 81.9 & 87.3 & 82.8 & 88.3\\
    \bottomrule
    \end{tabular}
    }
    \caption{Unsupervised phoneme segmentation results on English (TIMIT). ``L'' stands for the lenient metric commonly reported in the literature and ``H'' is the harsh metric proposed in \cite{Strgar2021-readout-seg}. $^*$ stands for results we obtained by running the code provided by the authors. ``Ours'' is a \unibitri/ ESPUM trained on TIMIT.}
    \label{tab:segment}
\end{table}

\begin{figure}[t]
    \centering
    \includegraphics[width=0.5\textwidth]{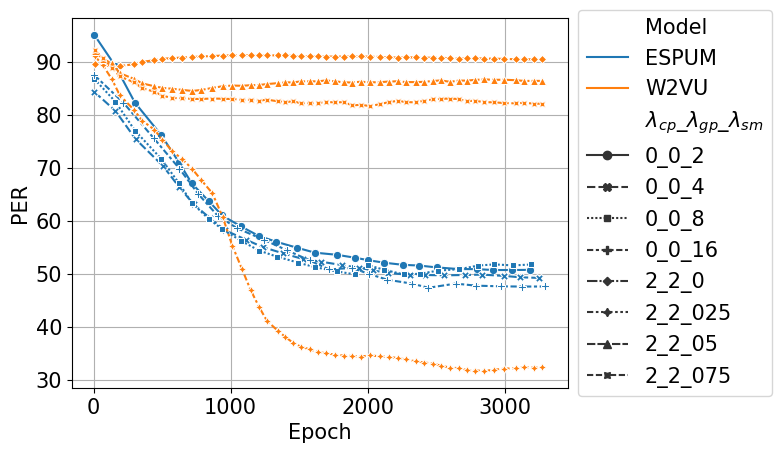}
    \caption{Convergence behavior of ESPUM (no segment relabeling) vs \wavtovecu/ over a range of hyperparameters defined in \cite{Baevski2021-wav2vec-u}, where $\lambda_{\mathrm{cp}}$ is the code penalty and $\lambda_{\mathrm{gp}}$ is the gradient penalty (neither used in ESPUM). $\lambda_{\mathrm{sm}}=\lambda_{\mathrm{smoothness}}$ is defined in \Eqref{eq:overall}.}
    \label{fig:roc_espum_vs_wav2vecu}
\end{figure}

\begin{table}[t]
    \centering
    \begin{tabular}{l|cc}
        \toprule
        Model & \begin{tabular}{@{}c@{}}
             PER\\
             (val)
        \end{tabular} & \begin{tabular}{@{}c@{}}
            Boundary F1\\
            (harsh)
        \end{tabular}\\
        \midrule
        bigrams only & 71.6 & 87.2 \\
       \unibi/ & 39.2 & 87.4\\
       \unibitri/ & \textbf{38.4} & 87.1 \\
       \unibifour/ & 40.0 & 86.5 \\
       uni+4-grams & 45.0 & 86.6 \\
       uni+5-grams & 77.9 & \textbf{87.6} \\
       uni+bi+tri+5-grams & 41.8 & 86.0 \\
        \bottomrule
    \end{tabular}
    \caption{Effect of different positional unigram and $N$-\skipgram/ combinations. All models use the segmentation from a \unibitri/ ESPUM after one segment relabeling iteration.} 
    \label{tab:eff_skipgram}
\end{table}

\begin{figure}[t]
    \centering
    \includegraphics[width=0.5\textwidth]{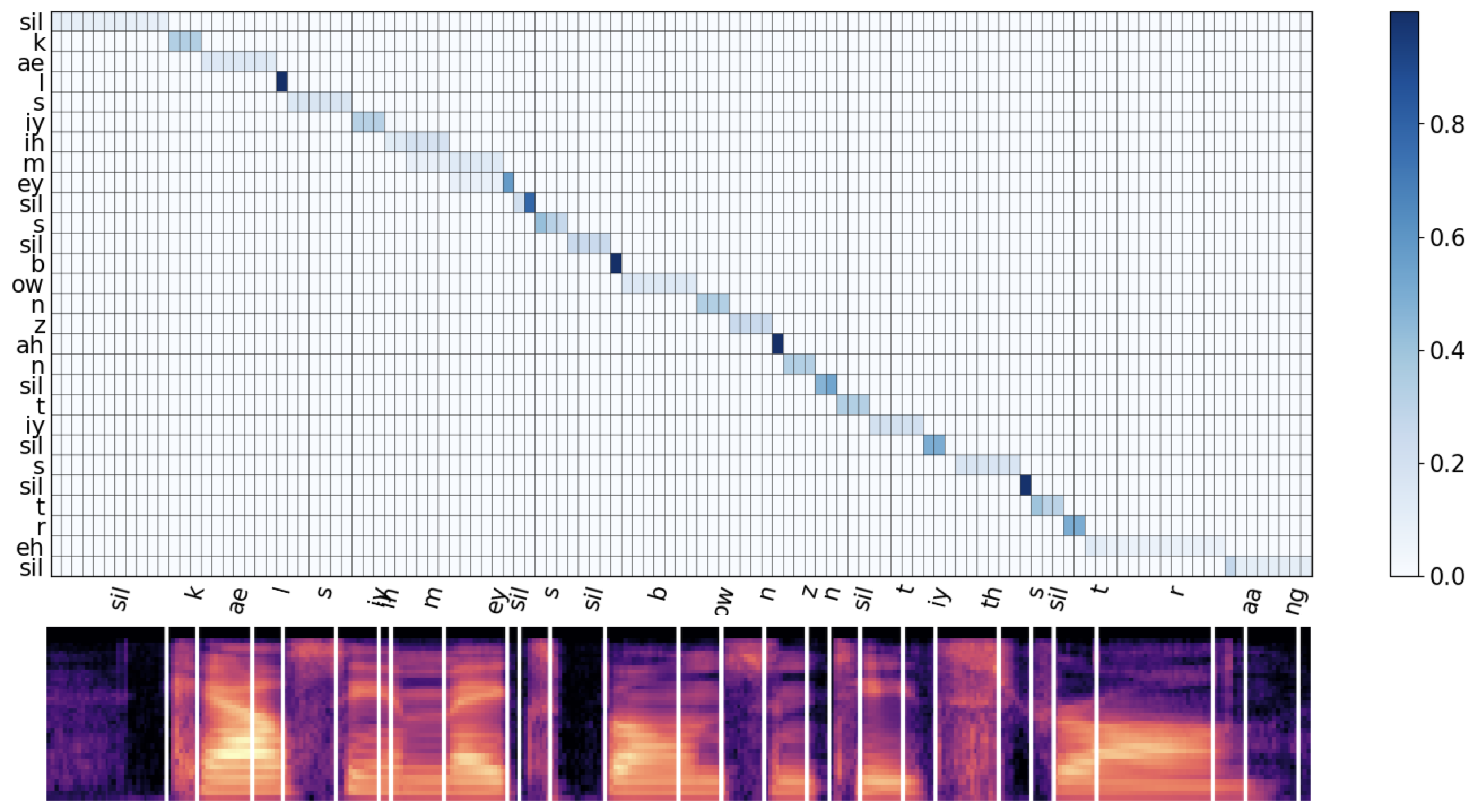}
    \caption{An example pooling matrix generated by a \unibitri/ ESPUM after one segment relabeling iteration.}
    \label{fig:pool_mat}
\end{figure}

\subsection{Experimental setup}
We use the TIMIT dataset~\cite{Garofolo1993} with the same split as in \cite{Yeh2019-asru,Baevski2021-wav2vec-u} for the \asru/ experiments. For the phoneme segmentation experiments, the full TIMIT test set is used instead to align with prior work ~\cite{Kreuk2020-cpc-seg}. We map the original 60 phonemes to 39 as in \cite{Baevski2021-wav2vec-u}.
Phoneme error rate (PER) is used to evaluate the ASR performance, while precision, recall, F1 and R-value metrics are used for the phoneme segmentation performance. For the latter task, we use both the \emph{lenient} scores in \cite{Yeh2019-asru,Kreuk2020-cpc-seg,Bhati2021-segmental-cpc} as well as the \emph{harsh} scores defined in \cite{Strgar2021-readout-seg} designed to avoid double counting of detected boundaries.

We use the 14-th layer of the \wavtovectwo/~\cite{Baevski2020-wav2vec2} model pretrained on 10,000-hour LibriLight~\cite{librilight} as the speech input features and a \kmeans/ module with 128 clusters for quantizing the speech features. The CNN generator is a one-layer CNN with a kernel size of $4$ and a stride size of $1$.
To obtain noisy phoneme boundary labels, we use the \readout/ model~\cite{Strgar2021-readout-seg} with the LibriLight \wavtovectwo/ encoder as the backbone. We then use the same 7-layer CNN in \cite{Strgar2021-readout-seg} as the CNN segmenter. During testing, we replace the soft alignment with a mean pooling within the predicted boundaries for decoding. We also found that training the segmenter using BCE loss with a positive weight of 1.1 and only labels with a confidence score above 0.6 from the \readout/ model achieves the best performance. We use $\lambda_{\mathrm{segment}}=1$ and $\lambda_{\mathrm{smoothness}}=16$ unless specified otherwise.

For all models, we only use bi-\skipgrams/ with skip sizes up to $6$, tri-\skipgram/s with skip sizes up to $2$ and the top 5000, or 68\% of the 4-grams and  50\% the 5-grams.
We conduct all our experiments on a 12GB \nvidia/ GPU. We implement our models using \fairseq/ and follow training settings in \cite{Baevski2021-wav2vec-u} if not specified. We train them end-to-end using \adam/ optimizer~\cite{Kingma2015-adam} with an initial learning rate of 0.004 and betas equal to $[0.5, 0.98]$ and a batch size of 640. We train the model further for one segment relabeling iterations for 20000 updates and observe further relabeling iteration leads to no significant improvement. We also experiment with HMM self-training (ST) techniques found previously to be effective~\cite{Chen2019-asru,Yeh2019-asru,Baevski2021-wav2vec-u}.

\subsection{Results}
The overall \asru/ result is shown in Table~\ref{tab:asru_timit}. Compared with EODM~\cite{Yeh2019-asru} before self-training, our model achieves a 8\% relative improvement in phone error rate (PER) in both the matched and unmatched setting. Further, segment relabeling of one iteration helps to reduce PER by 8.9\% relative. After self-training, while our model continues to outperform \cite{Yeh2019-asru} by 7.7\% relative PER in the matched setting, our model does not perform as well in the unmatched setting. This may be primarily due to the use of different language models used as well as discrepancy in hyperparameter settings of the self-training algorithms. Further, our model is still lagging behind the GAN-based \wavtovecu/~\cite{Baevski2021-wav2vec-u}.

The overall unsupervised phoneme segmentation result by our model is shown in Table~\ref{tab:segment}. With the help of unpaired text, our model trained in the matched setting  outperforms the best previous speech-only model~\cite{Strgar2021-readout-seg} by 5.4\% relative F1 and 4.2\% relative R-value, despite starting with segmentations with lower F1  (78.5\%) F1 due to discrepancy in training setting. It is also superior to EODM, the best speech+unpaired text models by 10.5\% relative F1 and 8.4\% relative R-value, though with the help of self-supervised representation pretrained on large speech corpora. We observe that segment relabeling helps to refine the segmentation by 2.3\% relative F1 and 2.9\% relative R-value. In addition, ESPUM consistently achieves 4.3\% relative F1 improvement over the speech-only segmenter~\cite{Strgar2021-readout-seg}, the teacher of our model in the unmatched setup, demonstrating the ability of our model to leverage unpaired textual information. Surprisingly, we found that \wavtovecu/+ST does not lead to good phoneme segmentation, trialing behind even the speech-only CPC-based approach~\cite{Kreuk2020-cpc-seg}, suggesting that our model is a superior forced aligner for low-resource languages.

\subsection{Analysis}
To compare the training stability of ESPUM and \wavtovecu/, we analyze their training convergence curves. As shown in \Figref{fig:roc_espum_vs_wav2vecu}, \wavtovecu/ fails to converge when $\lambda_{\mathrm{smoothness}}$ is too small or too large, while ESPUM remains consistent with a small improvement in PER as $\lambda_{\mathrm{smoothness}}$ increases from $0$ to $16$. Such improvement suggests that local homogeneity of phoneme distributions provide additional constraints to the generator besides those provided by the $N$-\skipgram/s matching constraints.

In addition to training stability, we also study the effect of distribution matching using different combinations of positional unigrams and $N$-\skipgrams/ in Table~\ref{tab:eff_skipgram}. We found that lower-order \skipgrams/ play bigger roles than higher-order \skipgrams/ for \asru/, evident by the fact that the \unibi/ model outperforms the uni+4/5-grams models. Also the information in the positional unigrams is crucial for \asru/ since the bigrams only model performs much worse than the \unibi/ model. While adding tri-\skipgrams/ help improve the PER of the \unibi/ model, adding 4-grams or 5-grams degrades its performance. Such diminishing return suggests that the information in the lower-order \skipgrams/ overlaps significantly with the higher-order $N$-grams. Although ~\cite{Yeh2019-asru} reported promising results using MFCC-based 5-gram matching, our version underperformed, possibly due to differences in feature extraction or segmentation methods.

In reviewing the pooling matrix from our segmenter in \Figref{fig:pool_mat}, we found that while most weights aligned with average pooling, the model sometimes leverages external context for phoneme prediction, as evidenced in certain phoneme examples .

\section{Conclusion}
In this work, we propose ESPUM, a novel GAN-free model for \asru/. On the standard benchmark TIMIT, ESPUM is more stable, memory-efficient and better-suited for unsupervised phoneme segmentation than previous \asru/ models. Future directions include making our model fully end-to-end and making it capable of higher-order distribution matching.

\vfill\pagebreak

\bibliographystyle{IEEEbib}
\bibliography{strings,refs}

\end{document}